# A Reconfigurable USAR Robot Designed for Traversing Complex 3D Terrain

*K. Davies, A. Ramirez-Serrano*

University of Calgary, Schulich School of Engineering, Dept. of Mechanical & Manufacturing Engineering,
kdavies@ucalgary.ca, aramirez@ucalgary.ca

## 1. Introduction

The use of robotics in Urban Search and Rescue (USAR) is growing steadily from their initial inception during the 2001 World Trade Centre incident. These first USAR robots were adapted from robots intended for tasks such as duct inspection and bomb disposal and proved useful for exploring voids (spaces within the collapsed structure) deemed too dangerous or too small for either canine or human searching. Physically, the original designs consisted of a solid robot body with two tracks, one to each side, which utilized either fixed or variable geometries as per the specific robot. Despite years of progress, the core design of robots currently in use for USAR purposes has deviated little, favoring software/control development and optimization of the basic robot template to improve performance instead. [1] [2]

Presented here is a novel design description of the *Cricket* (Fig. 1), an advanced robot with a broader range of capabilities than traditional USAR robots. By incorporating the tracked structure of earlier robots, appreciated for energy efficiency and robustness, into a multi-limbed walking design, the Cricket enables the use of advanced locomotion techniques. The ability to climb over obstacles many times the height of the robot, ascend vertical shafts without the assistance of a tether, and traverse rough and near vertical terrain improves the Cricket's capability to successfully locate victims in confined spaces.

## 2. Environmental Challenges

The most immediately apparent challenge presented by USAR operating environments comes from the terrain itself. The *Hot Zone* is a complex three dimensional environment littered with varying sizes of debris both fixed and free to move. Mission areas are almost exclusively enclosed voids with irregular sides presenting cross-sections ranging from meters across to several centimeters and often crisscrossed with collapsed structural elements and other debris. The route taken to explore these voids may be highly non-linear so as to avoid danger zones, pass through smaller openings in walls of debris, and move from one level of the structure to another. The robot may need to gain access to the mission zone by vertical decent through an opening cut into the structure and surfaces within the structure could present variable friction and stability properties. Performance by current USAR robots can be generally characterized as the largest surmountable obstacle for a given frontal area occupied by the robot. [2] [3]

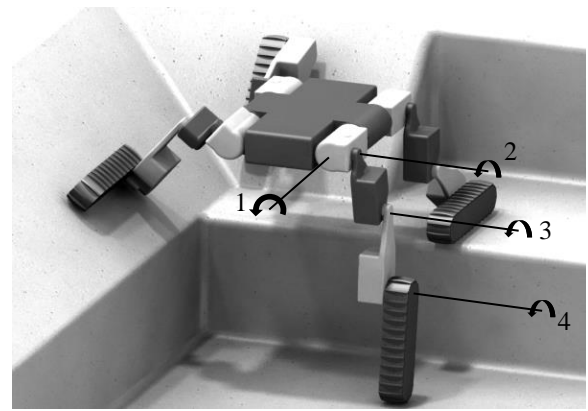

Fig. 1: Cricket Robot with Joint Indications

## 3. The Cricket Robot Design

The Cricket was created to address this performance criterion and allow for novel solutions to impasses observed in experience with current robots. By pairing a walking robot with a more conventional tracked design, the Cricket is capable of large variations in height thereby permitting passage through small openings and traversing obstacles several times the robot's minimal height. This is facilitated by the operating range of the articulated joints, 180° for the shoulder joint (Joint 1) and 360° for the three leg joints (Joints 2-4) as indicated in Fig. 1.

Incorporating an articulated walking system enables several advanced movements such as the ability to climb near-vertical surfaces in a fashion similar to a human rock climbing, provided that adequately sized openings within the surface are present. It is also possible to ascend smooth vertical shafts by pressing outwards against the walls of the shaft. Furthermore, as the Cricket was designed to be fully invertible and to be capable of moving forwards and backwards equally well, the need for complicated self-righting maneuvers, often requiring large spaces, is removed.

Physically, the Cricket is comprised of a solid body, housing the core electronics and power supply units, and four articulated multi-joint limbs, the last segment of which contains a tracked system similar to those found on more conventional USAR robots. Each joint is actuated independently via a stepper motor paired with a customized gearbox. Worm gears form the final drive in the joints to provide a large gear reduction for the volume of space occupied and reduce the amount of power required to lock the joints by creating a strong frictional resistance to inverse kinematics. Similarly, the track systems are actuated by DC motors paired with customized gearboxes utilizing worm gears for the final reduction. Finally, an IR ranger camera (SR 4000) is pivoted by means of a small servo allowing the robot to change viewing angles and create a 3D range map of the terrain. Combined with data acquired from the inertial measurement unit, GPS, and joint encoders, this is used for path planning and robot reconfiguration decisions.

### 4. Design Challenges

As a global strategy, the cricket was designed to be capable of supporting its own weight with only two limbs in any configuration, so as to allow the other two limbs to be repositioned freely. This resulted in significant strength demands both from the motors providing joint and track actuation as well as the frame structure. A secondary effect of the required torque output provided by the actuation mechanisms is a reduced movement speed. While it was originally planned that the Cricket would have a significantly higher ground speed than current USAR robots, the size/weight of the actuators needed proved unacceptable. Furthermore, as the motors increased in scale, the required power supply and structural support needed increased proportionally thereby mitigating the gains provided by larger motors due to increased weight. The final design of the Cricket provides a ground speed of 0.5 m/s which is comparable to faster USAR robots currently in use.

A second result of the complex locomotion system employed in the Cricket is the manufacturing complexity of structural elements. Due to external and internal loads applied to the structure, which can exceed 4 kN, the frame of the robot needs a high strength despite the small size of individual pieces. Furthermore, the high number of components associated with the actuation system combined with the overlap between leg segments restricts the volume within which structural elements can be placed. The final Cricket design was achieved both through the optimized selection of materials and by prioritizing stress reduction and strength over technical complexity. Iterative design based on finite element modeling was crucial in identifying areas which required strengthening or could be lightened, to reduce the overall weight of the robot, without reducing its adaptation/reconfiguration capabilities.

### 5. Conclusion

The Cricket presents a significant advancement in the capabilities of USAR robots by expanding the size and range of surmountable obstacles. Advanced movements, such as the scaling of near-vertical surfaces and the ascension vertical shafts without sacrificing reliability and operational lifespan, are made possible by combining a multi-joint walking system with the traditional tracked system. This increase in performance comes with an associated increase in robot complexity requiring a specialized purpose built structure.